# Persuasive Natural Language Generation - A Literature Review

Sebastian Duerr
Peter A. Gloor
MIT Center for Collective Intelligence
contact: {sduerr, pgloor}@mit.edu

This literature review focuses on the use of Natural Language Generation (NLG) to automatically detect and generate persuasive texts. Extending previous research on automatic identification of persuasion in text, we concentrate on generative aspects through conceptualizing determinants of persuasion in five business-focused categories: benevolence, linguistic appropriacy, logical argumentation, trustworthiness, tools & datasets. These allow NLG to increase an existing message's persuasiveness. Previous research illustrates key aspects in each of the above mentioned five categories. A research agenda to further study persuasive NLG is developed. The review includes analysis of seventy-seven articles, outlining the existing body of knowledge and showing the steady progress in this research field.

# 1. Introduction

The movie 'The Social Dilemma' by Jeff Orlowski (2020) explores the rise of social media and the damage it has caused to society. With a rather negative connotation, the directors address the topic of digital platforms and how their users are influenced and persuaded in surveillance capitalism (Economist 2019). Persuasion is an activity that involves one party, *the persuader*, trying to induce another party, *the persuadee*, to believe or disbelieve something or to do something (Iyer & Sycara 2019). The Economist (2019) claims that as a central tenet of surveillance capitalism, and persuasion is, furthermore, important in many aspects of daily life. Consider, for example, an employee demanding an increase in compensation, a physician trying to get a patient to enter a slimming programme, a charity volunteer trying to raise funds for a school project (Hunter et al. 2019), or a government advisor trying to get people to take a vaccination in the midst of a pandemic for the greater good.

A persuasive Natural Language Generation (NLG) artificial intelligence (AI) is a system that can create communications aimed at a user (the persuadee) in order to persuade her to accept a specific argument through persuasive messages. he persuadee benefits from eating vegetables to improve their health but is also confronted with opposing arguments to erase misunderstandings in the persuader's point of view. To do this, a persuasive NLG AI aims to use convincing arguments in order to persuade the persuadee. With recent advances in natural language processing and its subfield of natural language generation (NLG), it was demonstrated that pretrained language models (e.g., GPT-3) can achieve state-of-the-art results on a wide range of NLP tasks (Economist 2020). Such models allow for writing human-like texts through NLG, and can be fine-tuned for persuasion.

In the research of NLP and persuasion, Atalay et al. (2019) focus on syntax and persuasion, while Li et al. (2020) identify persuasion with NLP in online debates or in the news (Yu et al. 2019), and Rocklage et al. (2018) identify the relationship between psychological factors (e.g. emotions) and persuasion. In their seminal work, Iyer & Sycara (2019, p. 4) confer that "working with [subsequent uptake by the persuadee]" is an additional step. To explore this step, we conducted a structured literature review to identify whether the above research streams (natural language processing & persuasion) may fit in the following research question:

*What is the status quo of research focusing on persuasion and natural language generation?*

In this respect, a representative amount of research articles examined the concepts and technical intricacies behind persuasion in natural language processing and in psychological research. These aspects were classified and structured to develop a theory-based framework towards an overall understanding of persuasive NLG in a business context. Furthermore, we chose the format of a literature review for our paper, to indicate research gaps and survey an important part of larger research endeavors (vom Brocke et al. 2009).

## 2. Method

This paper's methodology follows a framework proposed by vom Brocke et al. (2009) which is based on a screening of the review literature itself and especially highlights the need for comprehensively documenting the process of literature search in such an article (Duerr et al. 2016). The framework is structured into the following five steps: (1) definition of review scope, (2) conceptualization of topic, (3) literature search, (4) literature analysis and synthesis, and (5) research agenda. Each of the steps will be briefly explained, when it will be addressed in the course of this work.

The first step is the definition of the review scope of this literature review. It is summarized in Figure 1 (categories applicable to this review on Persuasion and NLG research are highlighted) which is based on the taxonomy adapted by vom Brocke et al. (2009).

| Characteristic | Categories | | | |
|---|---|---|---|---|
| *Focus* | Research Outcome | Research Method | Theories | Applications |
| *Goal* | Integration | Criticism | | Central Issues |
| *Organization* | Historical | Conceptual | | Methodological |
| *Perspective* | Neutral Representation | | Espousal of Position | |
| *Audience* | Specialized Audience | General Scholars | Practitioners, Politicians | General Public |
| *Coverage* | exhaustive | exhaustive and selective | representative | central/pivotal |

*Figure 1. Taxonomy of this literature review on the collaborative use of Persuasion (following vom Brocke et al. 2009)*

This literature review *focuses* on outcomes of applied research in the domains of Persuasion and Natural Language Generation. The *goal* is to integrate findings with respect to five categories concluded from the business problem of persuading individuals through textual exchange (DeMers 2016). These categories were chosen as they address the psychological and technical aspects of persuasion and are a prerequisite to creating a persuasive NLG artificial intelligence. We selected this field because the artificial generation of persuasion through NLG has, as this literature review reveals, not been addressed in seminal articles following our structured research approach. Persuasion is already commonly studied in psychology (Quirk et al. 1985, Marwell & Schmitt 1967). Also, numerous studies in natural language processing focus on identifying and classifying persuasion in an automated way (Li et al. 2020, Rocklage et al. 2018, Iyer & Sycara 2019, Yu et al. 2019). This paper is *organized* along a *conceptual* structure. We did not take a particular *perspective* to provide a neutral representation of the results. As an *audience* of this review specialized scholars having an interest in the field of persuasion and the artificial generation of it were chosen. For *coverage*, our literature review can be categorized as representative as our research has been limited to certain journals, but does not consider the totality of the literature.

The second step is *conceptualization of the topic*.

It addresses the point that an author of a review article must begin with a topic in need of review, a broad conception of what is known about the topic, and potential areas where new knowledge may be needed. In the following, we conceptualize persuasion, and embed it into a business context. Furthermore, we introduce related theories, and conclude a categorization for the successive literature review.

In persuasion, the *persuader* induces a particular kind of mental state in the *persuadee*, e.g., through threats, but unlike an expression of sentiment, persuasion intends a change in the mental state of the persuadee (Iyer & Sycara 2019). Contemporary psychology and communication science (Rocklage et al. 2018, Park et al. 2015, Hunter et al. 2019) require the persuader to be acting intentionally, that is, the persuasive act. In the context of NLG, we usually refer to messages generated or augmented by an artificial intelligence, if we use the term persuasive act.

In his seminal work '*On Rhetoric*', Aristotle introduced his well-known ontology for persuasive acts. Accordingly, persuasion depends on multiple facets: emotions (*pathos*), logical structure of the argument (*logos*), the context (*cairos*), and on the speaker (*ethos*) (Schiappa & Nordin 2013).

Likewise, contemporary business literature conceptualizes persuasive acts through "principles of persuasion" (DeMers 2016). The author concludes that six interventions help at achieving persuasiveness. The first is being confident and remaining confident during the entirety of an appeal. Next, the introduction of logical argumentation fosters persuasiveness, since individuals are more inclined to be persuaded by logic. Additionally, making an appeal seem beneficial to the other party, by demonstrating the value of an appeal, choosing words carefully (i.e., selecting from a vocabulary that may be more persuasive), and using flattery (i.e., finding appropriate compliments) are recommended. Lastly, DeMers (2016) reveals that being patient and persistent (i.e., not to greatly alter one's approach) strengthens a persuader's persuasiveness. Next, we embed the presented "principles of persuasion" into related theories on persuasion (Cameron 2009).

Festinger's Theory of Cognitive Dissonance (1957) focuses on the relationships among cognitive elements, which include beliefs, opinions, attitudes, or knowledge (O'Keefe 1990). This relates most to Aristotle's *cairos*, and the need to create *benevolence* for the persuadee. Evaluating the 'business principles', this theory resonates best with what DeMers (2016) defines as 'making [the cognition] appealing to the other party'. However cognitions, and thus, a persuadee's perceived benevolence, can be dissonant, consonant, or irrelevant to each other. If a persuadee is presented with a sufficiently vast cognitive inconsistency, then they will perceive psychological discomfort, leading to an attempt to restore their cognitive balance through a reduction or elimination of the inconsistency (Stiff 1994, Harmon-Jones 2002). The magnitude of dissonance determines one's motivation to reduce it (Stiff 1994, Festinger 1957). Approaches towards reducing dissonance are: changing terms to make them more consonant, adding additional consonant percipience, or altering the magnitude of the percipience (Harmon-Jones 2002, Stiff 1994, O'Keefe 1990).

In 'principles of persuasion', DeMers (2016) contends that appropriate flattering and the usage of so-called high value words contribute to persuasive acts in business contexts (cf. Aristotle's *ethos*). Accordingly, Language Expectancy Theory (LET) identifies written or spoken language as a rule-based system through which persuadees develop expectations and preferences towards *"appropriate" linguistic* usage of words in varying situations (Burgoon & Miller 1985). Such expectations are frequently consistent with sociological and cultural norms, while preferences tend to relate to societal standards and cultural values (Burgoon & Miller 1985, Burgoon et al. 2002). Positive expectations that facilitate a persuasive act are, for instance, if a persuader stylizes a behavior that is perceived as more preferred than expected by the persuadee. In contrast, negative ones are inhibiting persuasion, e.g., when the persuader makes use of language that is considered to be socially unacceptable (Burgoon & Miller 1985, Burgoon et al. 2002).

Next, the 'principles of persuasion' confer that an argument based on logic is persuasive (DeMers 2016). What Aristotle terms *logos* is consistent with the theory of probabilistic models. Probabilistic models (McGuire 1981, Wyer 1970) are based on the rules of formal logic and probability, and predict beliefs regarding the conclusion of reasoning. These predictions are based on mathematical probability, and as such this theory is consistent with what Aristotle defines as *logos*. An exemplary belief syllogism (McGuire 1981) is composed of two premises that lead to a logical conclusion. The theorists (McGuire 1981, Wyer 1970) explain that believing in the premises leads to the expectation that the identified conclusion will follow. However, rather than solely thinking in all-or-nothing scenarios, beliefs can be ascertained through subjective probabilities: one's judgment of the probability that each of the premises is true (McGuire 1981, Wyer 1970). Furthermore, if a message evokes a perceptual change in the truth of the premise, or additional premises are supplemented, following this theory, a change in perceiving the conclusion is expected.

Last, Balance Theory focuses on the triadic relationship involving two individuals (e.g., persuader and persuadee), the persuadee's attitude toward the persuader (Aristotle's *pathos*), and their attitudes toward an attitudinal object (Heider 1958). The resulting triad can be balanced or unbalanced: This triad is in balance if all three relationships are positive, or one is positive and two are negative. If all three relationships are negative, or one is negative and two are positive, an unbalanced triad results, often motivating one to alter one of the three relationships (Heider 1958). Building on this theory, we aim to identify relational determinants that relate to improving the relationship between the persuader and the persuadee towards the attitudinal object (Heider 1958). In the business framework, we relate those determinants towards the pattern of *'trustworthiness',* that represent the persuadee's attitude towards the persuader. In his persuasion attempt, the persuader wants the persuadee to have a positive attitude.

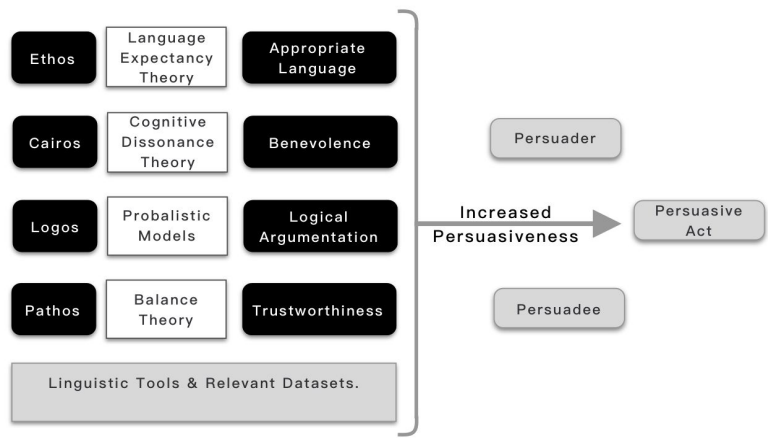

*Figure 2. Conceptualization of 'Business Principles'*

Figure 2 summarizes our conceptualization, starting from the business problem of being more persuasive, we adopt the 'principles of persuasion' (i.e., benevolence, linguistic appropriacy, logical argumentation, and trustworthiness), and embed them into different scholarly theories (i.e., balance theory, probabilistic models, theory of cognitive dissonance, and language expectation theory).

However, since we regard persuasion also from a technical perspective, (i.e., natural language generation), we also identify relevant data processing tools & datasets for persuasive natural language generation. In the following, we introduce our literature search process. Afterwards, we use these principles from managerial literature to categorize our identified aspects from our conducted literature review.

(3) The literature search considered the sources presented in Table 1

We searched on the journal quality evaluation web service 'simagojr.com' for the subjects 'psychology' and the subject category 'experimental and cognitive psychology'[1], respectively for the subject 'computer science' and the subject category 'artificial intelligence'[2] in December, 2020, to retrieve the most renowned journals in their respective research subject. We filtered the results for 'NAFTA', 'JOURNAL', and 'DECREASING SJR'. From the resulting list, the top 10 research journals in their domain were selected (see Table 1, Column 1). The relevant search terms used in the domains were: "Natural Language Processing (NLP)", "Natural Language Generation (NLG)", "Artificial Intelligence (AI)", "Persuasion", "Convincing", and "Negotiation". These evolved from readings related to our topic. We arrived at Table 1 by using the search string: 'source:"<JOUNRAL NAME>" (nlp OR nlg OR artificial intelligence) AND (persuasion OR persuade OR negotiation OR convincing)' on Google Scholar for each respective Journal.

| Journal | Domain | Search Field | Coverage | Hits | Relevant |
|---|---|---|---|---|---|
| Multivariate Behavioral Research | Psychology | Abstract | 2000-2020 | 2 | 0 |
| Journal of Memory and Language | Psychology | Abstract | 2000-2020 | 5 | 2 |
| Developmental Review | Psychology | Abstract | 2000-2020 | 11 | 0 |
| Cognitive Psychology | Psychology | Abstract | 2000-2020 | 94 | 6 |
| Journal of Experimental Psychology: General | Psychology | Abstract | 2000-2020 | 33 | 0 |
| Behavior Research Methods | Psychology | Abstract | 2000-2020 | 23 | 3 |
| Psychonomic Bulletin and Review | Psychology | Abstract | 2000-2020 | 1 | 0 |
| Journal of Experimental Psychology: Learning Memory and Cognition | Psychology | Abstract | 2000-2020 | 31 | 1 |
| Journal of Experimental Psychology: Human Perception and Performance | Psychology | Abstract | 2000-2020 | 26 | 3 |
| Cognitive Science | Psychology | Abstract | 2000-2020 | 57 | 2 |
| IEEE Transactions on Pattern Analysis and Machine Intelligence | Computer Science | Abstract | 2000-2020 | 93 | 1 |
| Foundations and Trends in Machine Learning | Computer Science | Abstract | 2000-2020 | 0 | 0 |
| IEEE Transactions on Neural Networks and Learning Systems | Computer Science | Abstract | 2000-2020 | 52 | 1 |
| IEEE Transactions on Fuzzy Systems | Computer Science | Abstract | 2000-2020 | 55 | 2 |
| Journal of Memory and Language | Computer Science | Abstract | 2000-2020 | 9 | 1 |
| Journal of Machine Learning Research | Computer Science | Abstract | 2000-2020 | 18 | 0 |
| Journal of the ACM | Computer Science | Abstract | 2000-2020 | 5 | 0 |
| International Journal of Intelligent Systems | Computer Science | Abstract | 2000-2020 | 74 | 4 |
| IEEE Transactions on Cognitive Communications and Networking | Computer Science | Abstract | 2000-2020 | 4 | 0 |
| Knowledge-Based Systems | Computer Science | Abstract | 2000-2020 | 251 | 3 |

*Table 1: Searched Journals on December 26, 2020*

Next, the choice of whether a retrieved article will be studied in detail in this literature review was made based on the abstract. After reading the identified articles and verifying their thematic consistency with the objective, the citations used in each article were analyzed to search for articles that have not been identified in the initial search process (Table 1).

---
[1] Journal of Experimental Child Psychology was deemed irrelevant and therefore not searched.
[2] Science Robotics, Int. Journal of Robotics Research was deemed irrelevant and therefore not searched.

Finally, Web of Science and Google Scholar are used to deploy forward and backward search (Webster & Watson 2002) for articles citing the identified article, and again it is analyzed whether these are consistent with the objective and have not been identified in the initial process (Duerr et al. 2016). This process served for enlarging the quantity of the main sources and unveiled another 48 relevant articles from journals, conferences, and magazines. The two last steps of von Brocke et al. (2009)'s framework for literature reviewing are (4) literature analysis and synthesis for classifying the identified articles as well as developing a (5) research agenda are explained thoroughly in the following sections, as these are the main outcomes of our work.

## 3. Results

The following paragraph shows the results of previous research (literature analysis and synthesis) and therefore is the next step (4) of the vom Brocke et al. (2009) literature review framework. Here, we focus on the four identified categories of persuasive natural language generation that underlie the business framework introduced in step (2) conceptualization. We ordered the identified categories alphabetically, hence, we do not imply a differentiation in degrees of persuasiveness. Additionally, we provide relevant tools and datasets required for implementation of a persuasive NLG.

## Benevolence

Determinants that aim at creating value for the persuadee are subsumed in this category (DeMers 2016, Voss & Raz 2016). In line with Cognitive Dissonance Theory (Festinger 1957) the identified eight determinants relate to altering a persuadee's perceived benevolence through dissonant or consonant measures (summarized in Table 3). An implementation in a persuasive NLG AI can be facilitated through identifying their absence or impact (Hunter et al 2019, Zarouali et al. 2020). The benevolence determinants are ordered alphabetically to not imply a specific order. The first column presents the determinants, the second column concisely defines each, and the third provides examples for all determinants that were identified. The last column states the corresponding citations.

| Determinant | Synopsis | Example | Source |
|---|---|---|---|
| Exemplifications | The process or act of giving examples. | Words such as: for instance, namely. | Quirk et al. 1985, Brewer 1980 |
| Appealing to Morality | Mentions of good or bad morality of a persuasive act. | A judge would sentence you since it is not okay to steal. | Marwell & Schmitt 1967, Luttrell et al. 2019 |
| Non-Monetary Terms | Offering of additional items that may be important to the persuadee but not to the persuader. | We do have an old reliable Toyota. We could just add this to the 5k. What do you think? | Ames & Wazlavek 2014 |
| None-Acceptable Terms | Understanding the persuadee's wants and thereby eliminating what is not. | Persuadee: 'This number is too low for me because I want to buy a car with it.' | Camp 2002 |
| Outcome | Mentioning of particular consequences from eventual actions. | I want you to put the gun down because I don't want to see you get hurt. | Douven & Mirabile 2018, Marwell & Schmitt 1967 |
| Regulatory Fit | Occurs when a message matches the persuadee's motivational orientation by focusing either on promoting gains or avoiding losses. | Example for a gain: 'This makes healthy teeth!', and a loss: 'This avoids cavities!" | Hirsh et al. 2012, Higgins 2000 |
| Scarcity | Mentioning of rarity, urgency, or opportunity for some outcome. | I tell you a little secret okay? They're pushing me to get something done and I am trying to hold them. | Cialdini & Goldstein 2002 |
| Social Proof | Reference to what is customary in a given situation. | Suppose your new car can be seen by all of your neighbors. | Cialdini & Goldstein 2002 |

*Table 3: Benevolence Determinants*

# Linguistic Appropriacy

This category subsumes fourteen determinants that facilitate an individual's stylome and aim at matching this with linguistic appropriacy. Such a stylome can be quantified and identified linguistically (Zarouali et al. 2020). Aforementioned Language Expectation Theory identifies written or spoken language as a rule-based system through which persuadees develop expectations (Burgoon & Miller 1985). The reason for profiling the stylome of an individual is to match these expectations (Park et al. 2015). Once implemented, a persuasive NLG AI can achieve congruence between a persuasive message and the persuadee, and thus generate persuasiveness. Table 2 introduces fourteen determinants of linguistic appropriacy in alphabetical order, provides a synopsis (i.e., brief summary, column two), an example for each determinant (column three), and the corresponding academic citation (in column four).

| Determinant | Synopsis | Example | Source |
|---|---|---|---|
| Amplifiers | These increase intensity, show precision or express certainty. | Words such as: extremely. | Quirk et al. 1985 |
| Connectivity | Degree to which a text contains explicit comparative connectives to express connections in it. | as ... as, more than ..., than ... | Crossley et al. 2008 |
| Downtoners | Reduce the strength of an expression or voice doubt. | Words such as: slightly, somewhat, almost. | Quirk et al. 1985 |
| Emphatics | Pronouns such as myself, yourself, herself, and himself. | Words such as: myself. | Quirk et al. 1985, Brewer 1980 |
| Evidence Words | Tendency to approve or disapprove something. | Words such as: according to. | Quirk et al. 1985 |
| Familiarity | Degree of familiarity of a word or how easily a word can be recognized by an adult. | Table, sun, and dog are more familiar than cortex or dogma. | Coltheart 1981, Hung & Gonzales 2013 |
| Hypernymy | Specificity or abstractness of a word. | Machine is a hypernym of a car. | Fellbaum 1998 |
| Imagability | Meaningful terms have a higher degree of meaning due to a semantic association with other words. | Words that are very imagery are bride or hammer, whereas quantum or dogmar are less. | Coltheart 1981, Nazari et al. 2019 |
| Indefinite Pronouns | Indefinite pronouns do not refer to a specific thing or person. | Words such as: all, non, some. | Quirk et al. 1985 |
| Lexical Overlap | Level to which phrases and words overlap in text and sentences. High overlap improves the cohesiveness and comprehension. | Possible overlaps between sentences: noun, argument, stem, and content word. | Kintsch & Van Dijk 1978 |
| Meaningfulness | Refers to the total number of varying words in a written text. | Words such as people are semantically related to many more compared to a noun such as abbess. | Wilson 1988, Jia 2009 |
| Open Ended Questions | Removal of aggression from a persuasive act that allows to introduce arguments without sounding dominant. | What else can I help you with? | Sinaceur & Tiedens 2006 |
| Temporal Cohesion | Consequent usage of one temporal tense (e.g., past or present). | He goes to school. Then, he goes home. (both are in presence) | McNamara et al. 2013 |
| Word Frequency | Indication of how often used words occur in a given language. | More uncommon words reflect that the writer possesses larger vocabulary. | Baayen et al. 1995 |

*Table 2: Linguistic Appropriacy Determinants*

## Logical Argumentation

Previous academic works unveil that arguments with consistent logic in persuasive acts increase persuasiveness (Cialdini & Goldstein 2002, Walton et al. 2008, Block et al. 2019). In line with the theory of Probabilistic Models (McGuire 1981, Wyer 1970), it is assumed that conclusive statements lead to a persuadee's expectation that a conclusion will follow. Technical implementations of logical argumentation or logical meaning representations occur as first order logic or semantic argumentation graphs (Moens 2018, Block et al. 2019). The first column of Table 4 enumerates our fourteen logical argumentation determinants, while the second provides a synopsis. Column three provides an example, and column four the corresponding citation in which the factor was identified. As in previous tables, the determinants are merely sorted alphabetically.

| Determinant | Synopsis | Example | Source |
|---|---|---|---|
| Analogy | Reframes issues through the usage of analogy or metaphor. | In the bible Moses saved all animals. Why don't you save those people? | Walton et al. 2008, Olguin et al. 2017 |
| Causal Cohesion | Related to causal relationships of actions and events that help to form relations between sentence clauses. | The ratio of causal verbs (e.g., break) to particles (e.g., because, due to). | Fellbaum 1998 |
| Connectives | Create explicit between clauses and sentences, and thus create cohesion between ideas. | E.g., 'moreover' or 'on the other hand'. | Longo 1994, Graesser et al. 2011 |
| Consistency | When references to previous commitments are made in order to persuade. | As I did this, you'll do that. | Cialdini & Goldstein 2002 |
| Establishing Ranges | Referencing to similar deals to establish the best possible trade-off range. | In the other deal, they agreed to pay only 5k but got a small car. Does that work for you? | Williams, 1983, Hyder et al. 2000 |
| Favors/Debts | When persuader implies that persuadee is indebted to him or her, e.g., coming from previous solicited or unsolicited favors. | So what do you say? Based on what we did for you, I think you should come outside. | Cialdini & Goldstein 2002, Britt & Larson et al. 2003 |
| Good/Bad Traits | Association of persuadee's mental states with good or bad traits. | Suppose you got a healthy body and a healthy mind, right? | Cialdini & Goldstein 2002, Bard et al. 2007 |
| Hedges | Express uncertainty or hesitation or to demonstrate indirectness. | Words such as: seem, tend, look, believe. | Tan et al. 2016 |
| Logical Operators | Establish explicit logical flow between concepts and describe the relation (e.g., 'if-then'). | Terms such as or, and, not, and if–then. | Costerman & Fayol 1997, Graesser et al. 2011 |
| Reason | Provides a justification for an argument based on additional arguments. | When people justify for actions or requests. | Walton et al. 2008, Fiedler & Horacek 2007, Corchado & Laza 2003 |
| Spatial Cohesion | Aids at constructing a spatial representation of text through the development of a situational model. | Location spatiality examples are beside, upon, here, and there; motion spatiality is represented through words like into and through. | Fellbaum 1998 |

*Table 4: Logical Argumentation Determinants*

## Trustworthiness

Trust plays an important role in the persuader-persuadee relationship. If established, the persuadee's attitude toward the persuader - as identified in Balance Theory (Heider 1958) - helps a persuadee to reason about the reciprocative nature, honesty or reliability of the counterpart (Kim & Duhachek 2020). An implementation of trustworthiness can, amongst others, be realized through identifying a persuadee's psychological profile (e.g., extroverted individuals respond better to texts that have a positive valence, and are in that case more persuadable, Zarouali et al. 2020, Park et al. 2015) to influence the degree of persuasiveness of a persuasive act. This category collates fifteen determinants pertaining to the increase or decrease of trustworthiness (column one, sorted alphabetically). The following columns provide a synopsis (column two), a corresponding example (column three) and the source in which the determinant was identified (column four).

| Determinant | Synopsis | Example | Source |
|---|---|---|---|
| Agreeableness | If a persuadee says "That's right," it indicates that he/she feels understood. | You want a car, is that right? | Van Swol et al. 2012, Fiedler & Horacek 2007 |
| Authority | Appealing or making reference to higher authority or experts to persuade. | We called your mom Mariam and she says please put the gun down and come outside. | Cialdini & Goldstein 2002, Catellani et al. 2020 |
| Seeking Comprehension | Instead of prioritizing own arguments, it is wise to focus on understanding the persuadee. | What do you mean by that? | Fisher Uri 1981, Kouzehgar et al. 2015 |
| Construal | Learning involves the generalization and abstraction from one's repeated experiences which is a high-construal mental process. | A short-term investor as opposed to long-term investor may rely more on a financial artificial intelligence. | Kim & Duhachek 2020, Abdallah et al. 2009 |
| Emotionality | The elicitation of positive or negative emotions to impose more weight on words. | Inclusion of words or expressions such as "amazing" or "excellent". | Rocklage et al. 2018 |
| Empathy | Attempts to connect with someone's emotional point of view. | Words and phrases like 'buddy' or 'friend'. | Cialdini & Goldstein 2002 |
| Labeling of Issues | Labeling counterpart's emotions after their identification, and verbalizing them for validation. | It feels like this situation is causing anxiety. | Lieberman et al. 2007 |
| Message-Person Congruence | Messages that are congruent with an individual's motivation are comprehended more easily and evaluated more positively. | In order to lose weight, we should eat less cheese. | Hirsh et al. 2012, Frey et al. 2019 |
| Personal Congruence | Crafting a message to fit the personality traits of the persuadee. | Since you are extroverted, I have this very exciting book with a happy end for you to read. | Zarouali et al. 2020, Douven & Mirabile 2018 |
| Politeness Marks | Make the hearer feel positive. | Words such as: I appreciate…, Nice work…. | Danescu-Niculescu-Mizil et al. 2013 |
| Repetition | Repeating the persuadee to encourage trust and familiarity. | Your last words were: I can do this? | Stephens et al. 2010 |
| Threat/Promise | Posing direct threats or promises. | Nobody will come in, but I want you to talk to me so we can help you. | Cialdini & Goldstein 2002, Walton et al. 2008, Zhou & Zenebe 2008 |
| User Beliefs | The beliefs of a persuadee that affect the likelihood that a persuasive act succeeds. | If I quit to smoke, I will get anxious about my studies, eat less, and lose too much weight. | Hunter et al. 2019, Hertz et al. 2016 |
| User Concerns | Some arguments may have a more pronounced impact on what a persuadee is concerned with. | If I quit on smoking, I will start to gain through eating more. | Hunter et al. 2019, Kaiser et al. 2011 |
| Valence | Positive or negative valence resonates differently with people, dependent on their psychological traits. | This is a very good positive book that will make you very happy. | Guerini et al. 2008b, Zarouali et al. 2020 |

*Table 5: Trustworthiness Determinants*

## Tools & Datasets

In the analyzed academic studies, we found that the authors use different datasets and tools to computationally process data for technical analyses of persuasion in NLP or NLG (e.g., in Guerini et al 2008a/b, Li et al. 2020, Iyer/Sycara 2019). Logically, the implementation of a persuasive NLG AI also depends on a variety of relevant tools and datasets which we identified and consolidated in Table 6. This table classifies our findings in types which are either tool or datasets (column one). We identified six tools and seventeen persuasion or message datasets. A software tool that is used in the context of persuasion and NLP, and datasets were chosen if they were used in the context of persuasion, textual exchange/debate and NLP. We further added a synopsis (column three) explaining every tool and dataset, providing a link (column four, if applicable) and the respective citation of the tool or dataset (column five). The tools and datasets are sorted alphabetically.

| Type | Name | Synopsis | Link | Citation |
|---|---|---|---|---|
| Tool | Args[dot]me | Argument resource search engine. | args.me | Ajjour et al. 2019 |
| Tool | Coh-Metrix | Provides an assortment of indices on the characteristics of words, sentences, and discourse. | http://cohmetrix.com/ | McNamara et al. 2013 |
| Tool | Evaluative Lexicon | Quantification of languages in terms of valence, extremity, and emotionality. | http://www.evaluativelexicon.com/ | Rocklage et al. 2018, Jia 2009 |
| Tool | Targer | Argument mining framework that is open sourced and can be used for tagging arguments in texts. | https://paperswithcode.com/paper/targer | Chernodub et al. 2019 |
| Tool | Textgain API | Conclusion of psychological traits based on words. | https://www.textgain.com/ | Zarouali et al. 2020 |
| Tool | Writing Pal | Artificial tutoring system providing writing strategy training. | http://www.adaptiveliteracy.com/writing-pal | McNamara et al. 2013, Reed & Grasso 2007 |
| Dataset | 16k Persuasiveness | 16k pairs of arguments over 32 topics annotated as to persuasiveness using crowdsourcing. | https://www.informatik.tu-darmstadt.de/ukp/research_6/data/argumentation_mining_1/ | Habernal & Gurevych 2016 |
| Dataset | Amazon Review Data | Database of approx. six million product reviews from amazon.com. | https://nijianmo.github.io/amazon/index.html | Jindal & Liu 2008 |
| Dataset | Args[dot]me Corpus | Comprises 387 606 arguments crawled from four debate portals in the middle of 2019. | https://webis.de/data/args-me-corpus.html | Ajjour et al. 2019 |
| Dataset | Argumentative Essay Dataset | Consists of about 402 essays with two files for each essay, the original and an annotated version. | https://www.informatik.tu-darmstadt.de/ukp/research_6/data/argumentation_mining_1/argument_annotated_essays_version_2/index.en.jsp | Eger et al. 2018 |
| Dataset | Blog Authorship | Corpus of 25,048 posts, of which around 457 were annotated with persuasive acts. | https://u.cs.biu.ac.il/~koppel/BlogCorpus.htm | Pranav Anand et al. 2011 |
| Dataset | ChangeMyView | Active community on Reddit that provides a platform where users present their own opinions and reasoning. | https://chenhaot.com/pages/changemyview.html | Tan et al. 2016, Yang et al. 2020 |
| Dataset | CORPs | Political speeches that are tagged with specific reactions such as APPLAUSE by an audience. | https://hlt-nlp.fbk.eu/corps | Guerini et al. 2008a |
| Dataset | DDO Corpus | Collection of approx. 80k debates from Oct' 2007 until Nov' 2017. | http://www.cs.cornell.edu/~esindurmus/ddo.html | Li et al. 2020 |
| Dataset | DebateSum | Approx. 188k evidence texts with extractive summaries and corresponding arguments. | https://mega.nz/folder/ZdQGmK6b#-0hoBWc5fLYuxQuH25feXg | Roush, Arvind Balaji 2020 |
| Dataset | Enron Sent Corpus | Contains 96,107 messages from the "Sent Mail" directories of all users. | https://wstyler.ucsd.edu/enronsent/ | Styler 2011 |
| Dataset | Penn Discourse Treebank | Database with 1 million annotated words of the WSJ corpus in Treebank-2. | https://www.seas.upenn.edu/~pdtb/ | Webber et al. 2019, Zhou & Zenebe 2008 |
| Dataset | Persuasion For Good Corpus | Collection of conversations generated by Mechanical Turk where the persuader tries to convince the persuadee to donate to charities. | https://convokit.cornell.edu/documentation/persuasionforgood.html | Wang et al. 2019 |
| Dataset | Persuasion Pairs | Contains textual pairs that consist of persuasive sentences and non-persuasive ones. | https://github.com/marcoguerini/paired_datasets_for_persuasion/releases/tag/v1.0 | Guerini et al. 2015 |
| Dataset | Pro/Con | Dataset with arguments on controversial issues shared by Procon.org. | https://github.com/marjanhs/stance | Hosseinia et al. 2019, Kabil & Eckbal 2020 |
| Dataset | Supreme Court Dialogs | Contains a collection of conversations from the U.S. Supreme Court Oral Arguments. | https://convokit.cornell.edu/documentation/supreme.html | Danescu-Niculescu-Mizil et al. 2013 |

*Table 6: Tools & Datasets - Overview*

# 4. Discussion

This section displays the last step of the framework for literature reviewing (Vom Brocke et al. 2009):
(5) Developing a research agenda.

For our proposed agenda for future research in the field of persuasive NLG (Figure 2), we conclude that an unambiguous and concise comprehension of how the forty-nine identified determinants influence the generation of persuasive artificially generated messages (i.e., the persuasive act) is needed. Furthermore, twenty-one tools & datasets were identified that allow one to train generative deep neural nets within the scope of persuasive NLG artificial intelligence. Next, we conclude our research proposals.

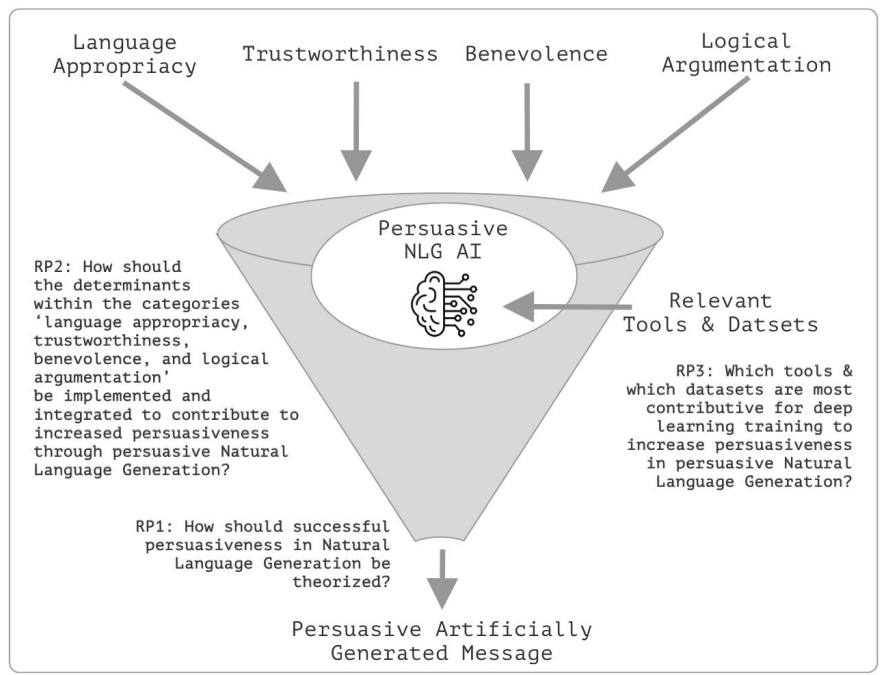

*Figure 3. Proposed research agenda for future research on persuasive NLG*

Our framework encompasses four identified categories (based on the 'principles of persuasion' *and* embedded into academic theories, i.e. Cognitive Dissonance, Language Expectancy, Probabilistic Models, and Balance Theory) as prerequisites for persuasive NLG that have not been comprehensively considered in the journals considered for this review before. For evaluating our approach the research proposal (RP1): *'How should successful persuasiveness in Natural Language Generation be theorized?'* has to be answered first in order to use this instrument as a remainder of further identified research proposals. Consequently, this framework can be used to investigate different *successful* approaches to generate persuasive messages through a persuasive NLG artificial intelligence.

Future research should investigate the empirical implementation of *benevolence for the persuadee*. In such regard, a given example in the circumstance of hostage negotiation (Gilbert 2010) may be transferable to business situations (cf. Table 3, Outcome: I want you to put the gun down because I don't want to see you get hurt). This example shows that the persuadee can expect benevolence as an outcome, if he acts in a certain way. Combining the identified determinants, a thorough linguistic analysis of *appropriate language* for the persuadee can be derived. As an example, meaningfulness is a crucial linguistic concept in persuasion (Atalay et al. 2019, Graesser et al. 2011; cf. Table 2), but lexical overlap even more directly influences persuasiveness, since it provides a consistency towards the persuadee's language expectancy (Heider 1958). A *consistent argumentative* logic implementable with determinants such as connectives, hedges, or logical operators, allows coherently concatenating a variety of arguments as well as the creation of an argumentative narrative. Logic can provide a blueprint for writing, or an approach to effectively organize a persuasive act (Habernal & Gurevych 2016). A high *trust-level* of a persuader would mean that s/he is likely to be chosen as an interaction partner (Axelrod 1984).

To the research community we propose a framework with persuasive determinants that are particularly pronounced in a persuasive act, which is also contingent on environmental aspects. Investigating these mechanisms (Table 3 - 6) would potentially provide insights regarding our research proposal (RP2): 'How should the determinants within the categories 'language appropriacy, trustworthiness, benevolence, and logical argumentation' be implemented and integrated to contribute to increased persuasiveness through persuasive Natural Language Generation?'

A variety of different tools and datasets prepare the input for deep learning models that underpin artificial intelligence and their training for text generation. Such models are inherently complex, so it is crucial to experiment, carefully prepare different datasets, and use the identified tools strategically to make the persuadee *act* upon the persuasive *act* (Anand et al. 2011). In this light, we propose RP 3: 'Which tools & which datasets are most contributive for deep learning training to increase persuasiveness in persuasive Natural Language Generation?'

Ultimately, the persuader will need to complement the persuasive measures that a persuasive NLG AI can suggest - due to possible deficiencies of crucial information that computational systems may inherently lack (e.g., aspects not explicitly outlined in textual data). Therefore, a persuasive NLG may be limited to serve as an assistant proposing suitable techniques or recommending alterations to linguistic measures in specific situations. Still, the persuader will be the one to edit and submit any artificially generated persuasive message to a persuadee, and is therefore *fully responsible*. Yet, such artificial intelligence can be used to help people to persuade them to do good things (like losing weight; Hunter et al. 2019).

## 5. Conclusion

This literature review synthesized the existing research on persuasive NLG in four categories for a persuasive NLG artificial intelligence by considering seventy-seven sources and integrating their results in forty-nine determinants. We concluded our identified categories and determinants (cf. Table 7) by addressing our previously introduced research question '*What is the status quo of research focusing on persuasion and natural language generation?*'

| Category | Determinants | # |
|---|---|---|
| Benevolence | Exemplifications, Appealing to Morality, Non-Monetary Terms, None-Acceptable Terms, Outcome, Regulatory Fit, Scarcity, and Social Proof. | 8 |
| Linguistic Appropriacy | Amplifiers, Connectivity, Downtoners, Emphatics, Evidence Words, Familiarity, Hypernymy, Imagability, Indefinite Pronouns, Lexical Overlap, Meaningfulness, Open Ended Questions, Temporal Cohesion, and Word Frequency. | 14 |
| Logical Argumentation | Analogy, Causal Cohesion, Connectives, Consistency, Establishing Ranges, Favors/Debts, Good/Bad Traits, Hedges, Logical Operators, Odd Numbers, Reason, and Spatial Cohesion. | 12 |
| Trustworthiness | Agreeableness, Authority, Seeking Comprehension, Construal, Emotionality, Empathy, Labeling of Issues, Message-Person Congruence, Personal Congruence, Politeness Marks, Repetition, Threat/Promise, User Beliefs, User Concerns, and Valence. | 15 |

*Table 7: Overview Findings of Literature Review*

Our findings provide an overview of the existing body of knowledge and propose a research agenda that unites and encompasses previous efforts. Previous research regarding persuasion and NLP has moved on from a strong persuasion identification focus, which supports our framework for generation. Identified articles were consistent but lacked sophisticated integration, therefore we see them in a rather fragmented stage. Additionally, we identified technical resources (tools & datasets) critical to AI success.

This literature review regarding persuasive NLG research faces some limitations itself. First, this literature review mainly covers the years 2000-2020. Undeniably, additional articles were published before, and in the meantime, which should be included in a future version. Secondly, this review concentrated only on a selection of top journals, but as we were not fully satisfied with our results we were using backward and forward search (Webster & Watson 2002), which may lead to less sophisticated sources. Moreover, it cannot be guaranteed that the framework will succeed or that it is complete. The presented approach identifies a vast amount of relevant aspects and should be used as a starting point for actions and for further research.

Therefore, it needs to be emphasized that no individual determinant suffices on its own. Rather multiple interactions in a given context will ensure a '*persuadee to be persuaded*' by a persuasive NLG that is built on the findings of this literature review. Unfortunately, some variables that influence a generated persuasive response cannot be deduced from text processing alone (e.g. current mental state, well-being, or the environment in which the persuadee receives the response), but should also be taken into consideration for persuasive acts (Hunter et al. 2019). In general, we do not see a 'one size fits all' approach (Duerr et al. 2016). Some linguistic determinants or persuasion techniques may work better than others in certain settings, but different in others. Pulse checks, data inputs, and reiterating the model, data and tools in the AI continuously, to learn from behaviors, attitudes, circumstances and usage will help. However, the identified articles, the detailed and transparent documentation of the literature search process, the proposed categorization of the aspects in each of the research fields, and the proposed research agenda can serve as a good starting point for further literature reviews and future research in the persuasive NLG research field.

To conclude, this paper has acknowledged that persuasive NLG builds on psychological, linguistic, and technical concepts. Despite the advantages of automated persuasion as presented in our introduction (e.g., entering a slimming programme, raising funds, taking vaccinations) with the help of AI, there is concern as to how such technologies can be misappropriated (cf. 'The Social Dilemma'). Ultimately, it is the academic community combined with advances of technological capabilities that will improve *persuasion for good* and release its great potential. Our research agenda suggests combining the right determinants in specific contexts and the usage of tools for training deep neural networks with relevant datasets. Thus, we have proposed a research direction to use the power of AI in this promising field for social good.

# References

Abdallah, S., D'souza, S., Gal, Y. A., Pasquier, P., & Rahwan, I. (2009). The effects of goal revelation on computer-mediated negotiation. In CogSci2009: Annual Meeting of the Cognitive Science Society (pp. 2614-2619). Cognitive Science Society.

Ajjour, Y., Wachsmuth, H., Kiesel, J., Potthast, M., Hagen, M., & Stein, B. (2019). Data Acquisition for Argument Search: The args.me Corpus. Lecture Notes in Computer Science (Including Subseries Lecture Notes in Artificial Intelligence and Lecture Notes in Bioinformatics), 11793 LNAI, 48–59. https://doi.org/10.1007/978-3-030-30179-8_4

Ames, D. R., & Wazlawek, A. S. (2014). Pushing in the dark: Causes and consequences of limited self-awareness for interpersonal assertiveness. Personality and Social Psychology Bulletin, 40(6), 775-790.

Atalay, A. S., Kihal, S. El, Ellsaesser, F., & Analytics, B. (2019). Using Natural Language Processing to Investigate the Role of Syntactic Structure in Persuasive Marketing Communication, (December), 1–57.

Axelord, R. (1984). The evolution of cooperation. New York.

Baayen, R. H., Piepenbrock, R., & Gulikers, L. (1995). CELEX. Philadelphia, PA: Linguistic Data Consortium.

Bard, E. G., Anderson, A. H., Chen, Y., Nicholson, H. B. M., Havard, C., & Dalzel-Job, S. (2007). Let's you do that: Sharing the cognitive burdens of dialogue. Journal of Memory and Language, 57(4), 616–641.

Block, K., Trumm, S., Sahitaj, P., Ollinger, S., & Bergmann, R. (2019). Clustering of Argument Graphs Using Semantic Similarity Measures. Lecture Notes in Computer Science (Including Subseries Lecture Notes in Artificial Intelligence and Lecture Notes in Bioinformatics), 11793 LNAI, 101–114. https://doi.org/10.1007/978-3-030-30179-8_8

Brewer, W. F. (1980). Literary theory, rhetoric, and stylistics: Implications for psychology. Theoretical Issues in Reading Comprehension: Perspectives from Cognitive Psychology, Linguistics, Artificial Intelligence, and Education, 221–239.

Britt, M. A., & Larson, A. A. (2003). Constructing representations of arguments. Journal of Memory and Language, 48(4), 794–810.

Miller, G. R. (2018). An expectancy interpretation of language and persuasion. In Recent advances in language, communication, and social psychology (pp. 199-229). Routledge.

Burgoon, M., Denning, V. P., & Roberts, L. (2002). Language expectancy theory. The persuasion handbook: Developments in theory and practice, 117-136.

Cameron, K. A. (2009). A practitioner's guide to persuasion: An overview of 15 selected persuasion theories, models and frameworks. Patient Education and Counseling, 74(3), 309–317. https://doi.org/10.1016/j.pec.2008.12.003

Camp, J. (2002). Start with no: The negotiating tools that the pros don't want you to know. Currency.

Catellani, P., Bertolotti, M., Vagni, M., & Pajardi, D. (2020). How expert witnesses' counterfactuals influence causal and responsibility attributions of mock jurors and expert judges. Applied Cognitive Psychology.

Chernodub, A., Oliynyk, O., Heidenreich, P., Bondarenko, A., Hagen, M., Biemann, C., & Panchenko, A. (2019, July). Targer: Neural argument mining at your fingertips. In Proceedings of the 57th Annual Meeting of the Association for Computational Linguistics: System Demonstrations (pp. 195-200).

Cialdini, R. B. (2001). Influence: Science and Practice. Book (Vol. 3rd).

Coltheart, M. (1981). The MRC psycholinguistic database. Quarterly Journal of Experimental Psychology, 33A, 497–505.

Corchado, J. M., & Laza, R. (2003). Constructing deliberative agents with case-based reasoning technology. International Journal of Intelligent Systems, 18(12), 1227–1241.

Costerman, J., & Fayol, M. (1997). Processing interclausal relationships: Studies in production and comprehension of text. Hillsdale, NJ: Lawrence Erlbaum Associates.

Crossley, S. A., Greenfield, J., & McNamara, D. S. (2008). Assessing text readability using cognitively based indices. TESOL Quarterly, 42, 475–493.

Danescu-Niculescu-Mizil, C., Lee, L., Pang, B., & Kleinberg, J. (2012, April). Echoes of power: Language effects and power differences in social interaction. In Proceedings of the 21st international conference on World Wide Web (pp. 699-708).

DeMers, J. (2016). "6 ways to persuade anyone of anything" in Business Insider July 2016; retrieved January 3, 2021.

Douven, I., & Mirabile, P. (2018). Best, second-best, and good-enough explanations: How they matter to reasoning. Journal of Experimental Psychology: Learning, Memory, and Cognition, 44(11), 1792.

Duerr, S., Oehlhorn, C., Maier, C., & Laumer, S. (2016). A literature review on enterprise social media collaboration in virtual teams: Challenges, determinants, implications and impacts. SIGMIS-CPR 2016 Proceedings of the 2016 ACM SIGMIS Conference on Computers and People Research, 113–122. https://doi.org/10.1145/2890602.2890611

Economist (2019): "Schumpeter Is Google an evil genius?" in *The Economist*: https://www.economist.com/business/2019/01/19/is-google-an-evil-genius; retrieved 26.12.2020.

Economist (2020): "A new AI language model generates poetry and prose" in *The Economist:* https://www.economist.com/science-and-technology/2020/08/06/a-new-ai-language-model-generates-poetry-and-prose; retrieved 26.12.2020.

Eger, S., Daxenberger, J., Stab, C., & Gurevych, I. (2018). Cross-lingual Argumentation Mining: Machine Translation (and a bit of Projection) is All You Need!. arXiv preprint arXiv:1807.08998.

Fellbaum, C. (1998). A semantic network of english: the mother of all WordNets. In EuroWordNet: A multilingual database with lexical semantic networks (pp. 137-148). Springer, Dordrecht.

Festinger, L. (1957). A theory of cognitive dissonance (Vol. 2). Stanford university press.

Fiedler, A., & Horacek, H. (2007). Argumentation within deductive reasoning. International Journal of Intelligent Systems, 22(1), 49–70.

Fisher, R., & Ury, W. (1981). FisherR. Getting to Yes: negotiating agreement without giving in.

Frey, S., Donnay, K., Helbing, D., Sumner, R. W., & Bos, M. W. (2019). The rippling dynamics of valenced messages in naturalistic youth chat. Behavior Research Methods, 51(4), 1737–1753.

Gilbert, I. V., & Henry, T. (2010). Persuasion detection in conversation. NAVAL POSTGRADUATE SCHOOL MONTEREY CA.

Graesser, A., McNamara, D. S., & Kulikowich, J. M. (2011). Coh-Metrix: Providing multilevel analyses of text characteristics. Educational Researcher, 40, 223–234.

Guerini, M., Strapparava, C., & Stock, O. (2008a). Valentino: A tool for valence shifting of natural language texts. Proceedings of the 6th International Conference on Language Resources and Evaluation, LREC 2008, 243–246.

Guerini, M., Strapparava, C., & Stock, O. (2008b). Resources for persuasion. Proceedings of the 6th International Conference on Language Resources and Evaluation, LREC 2008, 235–242.

Guerini, M., Özbal, G., & Strapparava, C. (2015). Echoes of persuasion: The effect of euphony in persuasive communication. arXiv preprint arXiv:1508.05817.

Habernal, I. & Gurevych, I. (2016). Which argument is more convincing? Analyzing and predicting convincingness of Web arguments using bidirectional LSTM. In Proceedings of the 54th Annual Meeting of the Association for Computational Linguistics (Volume 1: Long Papers). Pages: 1589-1599. Berlin, Germany. Association for Computational Linguistics.

Harmon-Jones, E. (2002). A cognitive dissonance theory perspective on persuasion. The persuasion handbook: Developments in theory and practice, 101.

Heider, F. (1958). The psychology of interpersonal relations Wiley. New York.

Hertz, U., Romand-Monnier, M., Kyriakopoulou, K., & Bahrami, B. (2016). Social influence protects collective decision making from equality bias. Journal of Experimental Psychology: Human Perception and Performance, 42(2), 164.

Higgins, E. T. (2000). Making a good decision: value from fit. American psychologist, 55(11), 1217.

Hirsh, J. B., Kang, S. K., & Bodenhausen, G. V. (2012). Personalized Persuasion: Tailoring Persuasive Appeals to Recipients' Personality Traits. Psychological Science, 23(6), 578–581. https://doi.org/10.1177/0956797611436349

Hosseinia, M., Dragut, E., & Mukherjee, A. (2019, July). Pro/Con: Neural Detection of Stance in Argumentative Opinions. In International Conference on Social Computing, Behavioral-Cultural Modeling and Prediction and Behavior Representation in Modeling and Simulation (pp. 21-30). Springer, Cham.

Hung, V. C., & Gonzalez, A. J. (2013). Context-Centric Speech-Based Human--Computer Interaction. International Journal of Intelligent Systems, 28(10), 1010–1037.

Hunter, A., Chalaguine, L., Czernuszenko, T., Hadoux, E., & Polberg, S. (2019). Towards Computational Persuasion via Natural Language Argumentation Dialogues. Lecture Notes in Computer Science (Including Subseries Lecture Notes in Artificial Intelligence and Lecture Notes in Bioinformatics), 11793 LNAI, 18–33. https://doi.org/10.1007/978-3-030-30179-8_2

Hyder, E. B., Prietula, M. J., & Weingart, L. R. (2000). Getting to best: Efficiency versus optimality in negotiation. Cognitive Science, 24(2), 169–204.

Iyer, R. R., & Sycara, K. (2019). An Unsupervised Domain-Independent Framework for Automated Detection of Persuasion Tactics in Text. ArXiv, 0(0), 1–19.

Jia, J. (2009). CSIEC: A computer assisted English learning chatbot based on textual knowledge and reasoning. Knowledge-Based Systems, 22(4), 249–255.

Jindal, N., & Liu, B. (2008, February). Opinion spam and analysis. In Proceedings of the 2008 international conference on web search and data mining (pp. 219-230).

Kapil, P., & Ekbal, A. (2020). A deep neural network based multi-task learning approach to hate speech detection. Knowledge-Based Systems, 210, 106458.

Kaiser, C., Schlick, S., & Bodendorf, F. (2011). Warning system for online market research--Identifying critical situations in online opinion formation. Knowledge-Based Systems, 24(6), 824–836.

Kim, T. W., & Duhachek, A. (2020). Artificial Intelligence and Persuasion: A Construal-Level Account. Psychological Science, 31(4), 363–380. https://doi.org/10.1177/0956797620904985

Kintsch, W., & van Dijk, T. (1978). Toward a model of text comprehension and production. Psychological Review, 85, 363– 394.

Kouzehgar, M., Badamchizadeh, M., & Feizi-Derakhshi, M.-R. (2015). Ant-Inspired Fuzzily Deceptive Robots. IEEE Transactions on Fuzzy Systems, 24(2), 374–387.

Li, J., Durmus, E., & Cardie, C. (2020). Exploring the Role of Argument Structure in Online Debate Persuasion. ArXiv, 8905–8912. https://doi.org/10.18653/v1/2020.emnlp-main.716

Lieberman, M. D., Eisenberger, N. I., Crockett, M. J., Tom, S. M., Pfeifer, J. H., & Way, B. M. (2007). Putting feelings into words. Psychological science, 18(5), 421-428.

Longo, B. (1994). Current research in technical communication: The role of metadiscourse in persuasion. Technical Communication,41, 348–352.

Luttrell, A., Philipp-Muller, A., & Petty, R. E. (2019). Challenging Moral Attitudes With Moral Messages. Psychological Science, 30(8), 1136–1150. https://doi.org/10.1177/0956797619854706

Marwell, G., & Schmitt, D. R. (1967). Dimensions of compliance-gaining behavior: An empirical analysis. Sociometry, 350-364.

McGuire, W. J. (1981). The probabilogical model of cognitive structure and attitude change. Cognitive responses in persuasion, 291-307.

McNamara, D. S., Crossley, S. A., & Roscoe, R. (2013). Natural language processing in an intelligent writing strategy tutoring system. Behavior Research Methods, 45(2), 499–515. https://doi.org/10.3758/s13428-012-0258-1

Moens, M. F. (2018). Argumentation mining: How can a machine acquire common sense and world knowledge? Argument and Computation, 9(1), 1–4. https://doi.org/10.3233/AAC-170025

Ye, K., Nazari, N. H., Hahn, J., Hussain, Z., Zhang, M., & Kovashka, A. (2019). Interpreting the Rhetoric of Visual Advertisements. IEEE Transactions on Pattern Analysis and Machine Intelligence.

O'Keefe, D. J. (1990). Persuasion: Theory of Research. Newbury Park, CA: Sage Publications.

Olguin, V., Trench, M., & Minervino, R. (2017). Attending to individual recipients' knowledge when generating persuasive analogies. Journal of Cognitive Psychology, 29(6), 755–768.

Park, G., Andrew Schwartz, H., Eichstaedt, J. C., Kern, M. L., Kosinski, M., Stillwell, D. J., … Seligman, M. E. P. (2015). Automatic personality assessment through social media language. Journal of Personality and Social Psychology, 108(6), 934–952. https://doi.org/10.1037/pspp0000020

Anand, P., King, J., Boyd-Graber, J. L., Wagner, E., Martell, C. H., Oard, D. W., & Resnik, P. (2011, January). Believe Me-We Can Do This! Annotating Persuasive Acts in Blog Text. In Computational Models of Natural Argument.

Quirk, R., Greenbaum, S., Leech, G., & Svartvik, J. (1985). A Comprehensive Grammar of the English Language Longman. London New York.

Reed, C., & Grasso, F. (2007). Recent advances in computational models of natural argument. International Journal of Intelligent Systems, 22(1), 1–15.

Rocklage, M. D., Rucker, D. D., & Nordgren, L. F. (2018). Persuasion, Emotion, and Language: The Intent to Persuade Transforms Language via Emotionality. Psychological Science, 29(5), 749–760. https://doi.org/10.1177/0956797617744797

Roush, A., & Balaji, A. (2020). DebateSum: A large-scale argument mining and summarization dataset. arXiv preprint arXiv:2011.07251.

Schiappa, E., & Nordin, J. P. (2013). Keeping faith with reason: A theory of practical reason.

Sinaceur, M., & Tiedens, L. Z. (2006). Get mad and get more than even: When and why anger expression is effective in negotiations. Journal of Experimental Social Psychology, 42(3), 314-322.

Stephens, G. J., Silbert, L. J., & Hasson, U. (2010). Speaker–listener neural coupling underlies successful communication. Proceedings of the National Academy of Sciences, 107(32), 14425-14430.

Stiff, J. B. (1994). The Guilford communication series.

Styler, Will (2011). The EnronSent Corpus. Technical Report 01-2011, University of Colorado at Boulder Institute of Cognitive Science, Boulder, CO.

Tan, C., Niculae, V., Danescu-Niculescu-Mizil, C., & Lee, L. (2016, April). Winning arguments: Interaction dynamics and persuasion strategies in good-faith online discussions. In Proceedings of the 25th international conference on world wide web (pp. 613-624).

Van Swol, L. M., Braun, M. T., & Malhotra, D. (2012). Evidence for the Pinocchio effect: Linguistic differences between lies, deception by omissions, and truths. Discourse Processes, 49(2), 79-106.

Brocke, J. V., Simons, A., Niehaves, B., Niehaves, B., Reimer, K., Plattfaut, R., & Cleven, A. (2009). Reconstructing the giant: On the importance of rigour in documenting the literature search process.

Voss, C., & Raz, T. (2016). Never split the difference: Negotiating as if your life depended on it. Random House.

D. Walton, C. Reed and F. Macagno, Argumentation schemes, Cambridge University Press, 2008.

Wang, X., Shi, W., Kim, R., Oh, Y., Yang, S., Zhang, J., & Yu, Z. (2019). Persuasion for good: Towards a personalized persuasive dialogue system for social good. arXiv preprint arXiv:1906.06725.

Webber, B., Prasad, R., Lee, A., & Joshi, A. (2019). The Penn Discourse Treebank 3.0 Annotation Manual.

Webster, J., & Watson, R. T. (2002). Analyzing the past to prepare for the future: Writing a literature review. MIS quarterly, xiii-xxiii.

Williams, G. R. (1983). Legal Negotiations and Settlement (St. Paul, Minnesota.

Wilson, M. D. (1988). The MRC psycholinguistic database: Machine readable dictionary, version 2. Behavioral Research Methods, Instruments, and Computers, 20, 6–11.

Wyer, R. S. (1970). Quantitative prediction of belief and opinion change: A further test of a subjective probability model. Journal of Personality and Social Psychology, 16(4), 559.

Yang, M., Huang, W., Tu, W., Qu, Q., Shen, Y., & Lei, K. (2020). Multitask Learning and Reinforcement Learning for Personalized Dialog Generation: An Empirical Study. IEEE Transactions on Neural Networks and Learning Systems.

Yu, S., Da San Martino, G., & Nakov, P. (2019). Experiments in detecting persuasion techniques in the news. ArXiv, 1–5.

Zarouali, B., Dobber, T., De Pauw, G., & de Vreese, C. (2020). Using a Personality-Profiling Algorithm to Investigate Political Microtargeting: Assessing the Persuasion Effects of Personality-Tailored Ads on Social Media. Communication Research. https://doi.org/10.1177/0093650220961965

Zhou, L., & Zenebe, A. (2008). Representation and reasoning under uncertainty in deception detection: A neuro-fuzzy approach. IEEE Transactions on Fuzzy Systems, 16(2), 442–454.